\documentclass[sigconf,nonacm, 10pt]{acmart}

\usepackage[english]{babel}

\usepackage{amsmath,amsfonts}
\usepackage{algorithmic}
\usepackage{graphicx}
 \usepackage{xcolor}

\usepackage{booktabs}
\usepackage[ruled,linesnumbered]{algorithm2e}

\usepackage{tikz}
\usetikzlibrary{shapes,arrows.meta,calc,decorations,arrows,positioning}

\usetikzlibrary{chains}
\usepackage{pgfplots}

\usetikzlibrary {patterns,patterns.meta}

\tikzstyle{nstyle}=[draw,circle, minimum size=12,fill=white,inner sep=0.2pt]%,circular drop shadow]
\tikzstyle{estyle}=[draw,line width=1 pt]
\tikzstyle{estyle-emph}=[draw,line width=2 pt,blue]
\tikzstyle{graphstyle}=[fill=gray!20]
\tikzstyle{smallnode}=[draw,circle, minimum size=1, inner sep=1]
\tikzstyle{snstyle}=[draw,circle, minimum size=2,fill=white,inner sep=0.1pt]
\tikzstyle{rootstyle}=[snstyle]
\tikzstyle{sestyle}=[draw]

\usepackage{comment}

\newtheorem{property}{Property}
\newcommand{\prior}{p}
\begin{document}

\title{Learning Filters with Certainty}
%\title{Learning Better with Filters using Beyond Binary Classification }
%Counting on Certainty: CBFs for Machine Learning
\author{Yuval Banoun}
\affiliation{%
  \institution{Technion}
  \country{Israel}
}

\author{Daniel Sadoc Menasché}
\affiliation{%
  \institution{UFRJ}
  \country{Brazil}
}

\author{Ori Rottenstreich}
\affiliation{%
  \institution{Technion}
  \country{Israel}
}

\begin{abstract}
Hash-based data structures such as Bloom filters are widely used in network systems for tasks including caching, anomaly detection, and machine learning pipelines. They typically provide binary indications of whether an element belongs to a set of interest, e.g., the contents of a cache. When uncertainty arises due to hash collisions, a positive indication is returned to avoid false negatives. We argue that the certainty associated with such indications can itself be useful information. This work focuses on Counting Bloom Filters (CBFs), a Bloom-filter variant that maintains counters rather than bits. Besides supporting insertions and deletions, these counters provide additional information that can be used to estimate the certainty of positive membership indications. We show how this certainty signal can be exploited in architectures that combine Bloom Filters with machine learning (ML) models. 
% The counter values maintained by CBFs enable estimation of the correctness probability of positive membership indications. 
\end{abstract}

\maketitle

\section{Introduction}
\thispagestyle{empty}

\emph{Sketches} often represent data structures and algorithms used for summarizing massive data sets while enabling answering queries on the data~\cite{Bloom, Cormode17}. Typically the sketch allocated memory is much smaller than the memory of the original data (e.g., logarithmic in its size). Accordingly, an error is permitted in the query answers and a tradeoff exists between the accuracy and memory efficiency. % Often a sketch is designed to answer a set of   queries with regards to the data.

The Bloom filter (BF)~\cite{Bloom} is a popular data structure   used in many networking device algorithms,
in fields as diverse as packet classification, routing, filtering, caching, and blockchain networks~\cite{BloomNet, SoK24, Voronov24}, as well as beyond networking, e.g., in  spell checking~\cite{mullin1990tale},  verification~\cite{Bloomformal04} and ML pipelines~\cite{dai2020adaptive, santiago2020weightless,kraska2018case, LBF, lin2025ensemble}. The BF is used for set representation, supporting element insertion, and answering membership queries. There are two kinds of errors in membership queries: a false positive (when an element $x \notin S$ is reported as a member of a  set $S$) and a false negative (when  $x \in S$ is reported as a non-member of $S$). The BF encounters false positives and has no false negatives.   

The BF is built as an array of bits, where hash functions are used to map elements to locations in the array. 
With initial values of zero bits, the elements of $S$ are   inserted to the filter, setting all bits pointed by the hash functions. Upon a query, the bits mapped by the queried element are examined and a positive answer is returned   when the bits are all set.

Seminal papers by Kraska et al.~\cite{kraska2018case} and by Mitzenmacher~\cite{LBF} presented how BFs can interact well in multiple forms with machine learning (ML) models such as:  (i) Using a learning model as a pre-filter before the Bloom filter. If the score $f(x)$  for an element $x$ is at least a threshold $\tau$, indicate membership. The Bloom filter is queried only when the score $f(x)$ reported by the model is low to avoid potential false negatives of the model. (ii) Using a Bloom filter before the learning model to remove most queries for non-member elements and reduce the amount of false positives. 

To avoid false negatives, Bloom filters do not support deletion of elements from the represented set. 
Deletions are necessary whenever the set is dynamic.
%Deletions are necessary whenever the set if dynamic. 
Counting Bloom Filters (CBFs)~\cite{CBF}  use counters in the Bloom filter structure, thus also allowing for deletions within counter limits. This requires overhead in memory as often a counter is represented as multiple bits (typically there are four bits per counter). Simply, a membership query is answered as positive if the corresponding counters are not zero. 

In~\cite{BloomParadox} the authors uncovered an   additional capability of     CBFs:  \emph{Estimating the correctness probability of a positive membership indication}. This allows using a CBF as an engineering tool that provides a continuous confidence score rather than a binary yes/no answer to a membership query.

Motivated by the growing use of Bloom filters in machine learning (ML) systems, our work explores new applications of such ability to estimate the correctness probability of query answers. We view the memory overhead introduced by the counters in CBFs not only as a mechanism for supporting deletions, but also as an information-rich signal that can quantify the certainty of membership queries. In particular, the counter values allow an ML model to reason about uncertainty in a way that a standard Bloom filter cannot.

Our main contributions are as follows.
First, building on the observation of~\cite{BloomParadox} that Counting Bloom Filters (CBFs) can estimate posterior membership probabilities from their counters, we view CBFs as certainty-aware data structures whose outputs can serve as confidence signals. 
Second, we propose four architectures that combine learned membership models with Bloom filters and CBFs. For each architecture, we describe the algorithm and analyze its false-positive rate, discussing the tradeoffs among these architectures in terms of false positives, inference cost, memory footprint, and the benefits of certainty-aware decisions.

    % \item We discuss dynamic-update issues that arise when combining learned models with mutable set representations, and outline practical deployment considerations for such hybrid systems.

The remainder of this paper is organized as follows. In Section~\ref{sec:background}, we review Counting Bloom Filters and in Section~\ref{sec:traditional_architectures}, we summarize the traditional learned Bloom filter and sandwiched learned Bloom filter architectures. In Section~\ref{sec:certainty_architectures}, we introduce four certainty-aware architectures that leverage CBF counter values, and we analyze their false-positive rates and design tradeoffs. Finally, Section~\ref{sec:dynamic_updates} concludes the paper with a discussion of practical deployment considerations and future research directions. 

\section{Background and Related Work}
\label{sec:background}
A Counting Bloom Filter (CBF)
replaces the array of bits in a Bloom filter with an array of  counters~\cite{CBF}. Each element is mapped by $k$ hash functions  to counters. Insertions and deletions are supported by incrementing or decrementing the counters, respectively.   Upon a membership query, a positive indication is returned when all $k$ counters are positive. 
 
Consider a set $S \subseteq U$ represented by a CBF. 
Let $V(x) = (c_1,  \ldots, c_k)$ be 
the values of the $k$ counters to which an element $x$ maps.  
Let $P(x \in S)$ be the prior membership probability of the element $x$ in the set $S$ before accessing the filter. This probability might be uniform over all elements in the universe (and thus equal $|S|$ / $|U|$ if the universe  $U$ is finite) or 
might vary across elements, so that some elements have higher prior membership probability than others. The membership probability based on the counters equals~\cite{BloomParadox} 
\begin{align}
&P(x \in S \mid V(x))  \nonumber\\
&\quad=
\frac{m^k \cdot \left(\prod_{j=1}^{k} c_j\right) \cdot P(x \in S)}
{m^k \cdot \left(\prod_{j=1}^{k} c_j\right) \cdot P(x \in S)
+ (n \cdot k)^k \cdot \left(1 - P(x \in S)\right)},
\label{eq:cbf_posterior_base}
\end{align}
where $k$ is the number of hash functions, $n$ is the set size, and $m$ is the number of counters in the CBF.

In~\cite{BloomParadox}, the authors suggested the following policy for a decision when the costs of both false negative and false positive are finite such that the cost of a false negative equals $\alpha$ times that of a false positive. 

\begin{property}
An optimal decision policy for the CBF is to declare membership if $P(x \in S \mid V(x)) \ge {1}/({\alpha+1})$.
\end{property}

We illustrate the probability through the following simple experiment. Consider a set with $n=1000$ elements drawn from among a universe of size 10000. The set is represented in a CBF with $m$=5000 counters using $k=3$ hash functions. When set elements are selected uniformly from the universe, the prior membership probability of each element $x$ equals $P(x \in S)$ = 1000/10000 = 0.1. We repeated that experiment 500 times and in each we examined the values of the CBF counters for the 1000 elements in the set and for the 10000-1000=9000 elements outside of the set. 

\begin{figure}[t]
    \centering
    \begin{minipage}[t]{0.51\textwidth}
        \centering
\begin{tikzpicture}
\begin{axis}
[ 
    xlabel={product of counters $\Pi$},
    ylabel = {probability},
    xmin=1, xmax=20,
    ymin = 0, ymax=0.5,
    xtick = {1, 2, 3, 4, 5, 6, 7,8, 9,10, 11, 12, 13, 14, 15, 16, 17, 18, 19, 20}, 
    ymajorgrids,
    ytick={0.0,0.1, 0.2, 0.3, 0.4, 0.5, 0.6, 0.7, 0.8, 0.9, 1.0},
    height = 0.503\textwidth,
    width = 0.87\textwidth, 
    legend style={nodes={scale=0.60}, at={(0.845,0.853)}}, 
    tick label style={font=\footnotesize} 
] 
\addplot[color=blue, mark=triangle]
coordinates{
(1, 0.164932)
(2, 0.29797)
(3, 0.08962)
(4, 0.195694)
(5, 0.002542)
(6, 0.106806)
(7, 3.6e-05)
(8, 0.057628)
(9, 0.016208)
(10, 0.003078)
(11, 0.0)
(12, 0.039288)
(13, 0.0)
(14, 5.4e-05)
(15, 0.00102)
(16, 0.007258)
(17, 0.0)
(18, 0.009786)
(19, 0.0)
(20, 0.001136)
};
\addlegendentry{Members};
\addplot[color=red, mark=o]
coordinates{
(1, 0.3883375705398586)
(2, 0.3499691839525433)
(3, 0.07024132817164538)
(4, 0.11508349222857803)
(5, 0.0012206043797307448)
(6, 0.04234317520848982)
(7, 1.2037518537778549e-05)
(8, 0.016876600989965525)
(9, 0.004311839140232276)
(10, 0.0007198436085591571)
(11, 0.0)
(12, 0.007593266693630708)
(13, 0.0)
(14, 7.222511122667129e-06)
(15, 0.00013722771133067545)
(16, 0.001093006683230292)
(17, 0.0)
(18, 0.0012928294909574161)
(19, 0.0)
(20, 0.00010593016313245123)
};
\addlegendentry{Non-members};
\end{axis}
\end{tikzpicture}
        \caption*{(a) Distribution of product of counters $\Pi$ for elements with a positive indication}
        \label{fig:product_distribution}
    \end{minipage}    
    \begin{minipage}[t]{0.51\textwidth}
        \centering
\label{figure_membership_probability_vs_counter_product}
\begin{tikzpicture}
\begin{axis}
[ 
    xlabel={product of counters $\Pi$},
    ylabel = {membership probability},
    xmin=1, xmax=20,
    ymin = 0, ymax=1,
    xtick = {1, 2, 3, 4, 5, 6, 7,8, 9,10, 11, 12, 13, 14, 15, 16, 17, 18, 19, 20}, 
    ymajorgrids,
    ytick={0.0,0.1, 0.2, 0.3, 0.4, 0.5, 0.6, 0.7, 0.8, 0.9, 1.0},
    height = 0.503\textwidth,
    width = 0.87\textwidth, 
    legend style={nodes={scale=0.60}, at={(0.725,0.353)}}, 
    tick label style={font=\footnotesize} 
] 
\addplot[color=red, mark=triangle, only marks]
coordinates{
(1, 0.339673913)
(2, 0.5070993915)
(3, 0.6067961165)
(4, 0.6729475101)
(5, 0.7200460829)
(6, 0.7552870091)
(7, 0.782647585)
(8, 0.8045052293)
(9, 0.8223684211)
(10, 0.8372404555)
(11, 0.8498145859)
(12, 0.8605851979)
(13, 0.8699143469)
(14, 0.8780732564)
(15, 0.8852691218)
(16, 0.8916629514)
(17, 0.8973817568)
(18, 0.9025270758)
(19, 0.9071810542)
(20, 0.911410864)
};
\addlegendentry{Theory};
\addplot[color=black, mark=o, only marks]
coordinates{
(1, 0.33829568156738554)
(2, 0.5061474226348814)
(3, 0.6056551239423674)
(4, 0.6718000123584783)
(5, 0.7148481439820022)
(6, 0.7522502852474258)
(7, 0.782608695652174)
(8, 0.8043211255024565)
(9, 0.81899949469429)
(10, 0.8373231773667029)
(12, 0.8616545310992192)
(14, 0.9)
(15, 0.8994708994708994)
(16, 0.8888072495713936)
(18, 0.9011049723756906)
(20, 0.9281045751633987)
};
\addlegendentry{Simulation};
\end{axis}
\end{tikzpicture}
        \caption*{(b) Membership probability (with prior membership probability $P(x \in S)=0.1$)}

    \end{minipage}

    \begin{minipage}[t]{0.51\textwidth}
        \centering
\label{figure_membership_probability_vs_counter_product_below}
\begin{tikzpicture}
\begin{axis}
[ 
    xlabel={product of counters $\Pi$},
    ylabel = {membership probability},
    xmin=1, xmax=20,
    ymin = 0, ymax=1,
    xtick = {1, 2, 3, 4, 5, 6, 7,8, 9,10, 11, 12, 13, 14, 15, 16, 17, 18, 19, 20}, 
    ymajorgrids,
    ytick={0.0,0.1, 0.2, 0.3, 0.4, 0.5, 0.6, 0.7, 0.8, 0.9, 1.0},
    height = 0.503\textwidth,
    width = 0.87\textwidth, 
    legend style={nodes={scale=0.60}, at={(0.845,0.403)}}, 
    tick label style={font=\footnotesize} 
] 
\addplot[color=orange, mark=o, only marks]
coordinates{
(1, 0.6067961165)
(2, 0.7552870091)
(3, 0.8223684211)
(4, 0.8605851979)
(5, 0.8852691218)
(6, 0.9025270758)
(7, 0.9152719665)
(8, 0.9250693802)
(9, 0.9328358209)
(10, 0.9391435011)
(11, 0.9443681319)
(12, 0.9487666034)
(13, 0.9525205158)
(14, 0.9557618788)
(15, 0.9585889571)
(16, 0.9610764056)
(17, 0.9632819583)
(18, 0.9652509653)
(19, 0.967019544)
(20, 0.9686168152)
};
\addlegendentry{Theory $P(x \in S)=0.3$};
\addplot[color=red, mark=triangle, only marks]
coordinates{
(1, 0.339673913)
(2, 0.5070993915)
(3, 0.6067961165)
(4, 0.6729475101)
(5, 0.7200460829)
(6, 0.7552870091)
(7, 0.782647585)
(8, 0.8045052293)
(9, 0.8223684211)
(10, 0.8372404555)
(11, 0.8498145859)
(12, 0.8605851979)
(13, 0.8699143469)
(14, 0.8780732564)
(15, 0.8852691218)
(16, 0.8916629514)
(17, 0.8973817568)
(18, 0.9025270758)
(19, 0.9071810542)
(20, 0.911410864)
};
\addlegendentry{Theory $P(x \in S)=0.1$};
\addplot[color=green, mark=star, only marks]
coordinates{
(1, 0.1336898396)
(2, 0.2358490566)
(3, 0.3164556962)
(4, 0.3816793893)
(5, 0.4355400697)
(6, 0.4807692308)
(7, 0.5192878338)
(8, 0.5524861878)
(9, 0.5813953488)
(10, 0.6067961165)
(11, 0.6292906178)
(12, 0.6493506494)
(13, 0.6673511294)
(14, 0.68359375)
(15, 0.6983240223)
(16, 0.7117437722)
(17, 0.7240204429)
(18, 0.7352941176)
(19, 0.7456828885)
(20, 0.7552870091)
};
\addlegendentry{Theory $P(x \in S)=0.03$};
\end{axis}
\end{tikzpicture}
        \caption*{(c) Membership probability for different prior membership probabilities}

    \end{minipage}
\caption{Probability of membership based on the product of counter values in the CBF. }
\label{figure_counter_product_stats}
\end{figure}

Fig.~\ref{figure_counter_product_stats}(a) shows the 
distribution of the product of counters  for all those with a positive indication. Accordingly, the product is 1 or above. The two curves refer to members and non-members. For non-members, the value of 1 is the most common and is observed for 0.3883 of the elements. 
For members, it appears with probability 0.1649, and the most common value is 2. Values of 4 and above appear   
with probability 0.1915 for non-members  and  
with probability 0.4475 for members. 

Fig.~\ref{figure_counter_product_stats}(b) shows the probability of membership based on the counter values, comparing the simulation results with the above formula of $P(x \in S \mid V(x))$. It demonstrates the variability in certainty, namely the probability for the correctness of a positive indication. While it equals 0.3397 when the product equals 1 (namely all counters have value of 1), the probability is 49.2\% higher with a value of 0.5071 for a product of 2, upon a change in a single counter value to 2. 
For a product of 8, the probability exceeds 0.8 and reaches 0.8605 when the product is 12.
%For product of 8 the probability reaches a value of over 0.8000  and reaches 0.8605 when the product is 12.
We see very high similarity of the theory and simulation probabilities for small values of the product. As higher product values are rare, we see a small difference between the two values. \footnote{Note that product values which are above 10 and are of prime values require having a single counter with such high value. As it happens with very small probability, all such cases have not been observed even once in the simulation so   the corresponding curve contains fewer points.}

Fig.~\ref{figure_counter_product_stats}(c) shows the impact of the prior membership probability on $P(x \in S \mid V(x))$ based on its formula from above. For simplicity, we use the same parameter values $k, n$ and $m$ but consider $P(x \in S) \in \{0.03, 0.1, 0.3\}$.  $P(x \in S \mid V(x))$ 
is highly influenced by the prior probability 
and for any value of the product increases monotonically by that value. With product of 1, the probability equals $P(x \in S)$ 
0.1337,  0.3396, 0.6067 for the prior probability values of  0.03, 0.1 and 0.3. When the product is 4, the probabilities are   0.3817, 0.6729 and 0.8606.
When the product is 8, the probabilities are as high as 0.5525, 0.8045 and 0.9251. This demonstrates the high impact of the counter values, especially in the case of low prior probability. For 
$P(x \in S)=0.03$, the $P(x \in S \mid V(x))$ is 2.85x and 4.13x higher when the product is 4 and 8 than when it equals 1.
If $P(x \in S)=0.3$, the increase is smaller: $P(x \in S \mid V(x))$ is 1.42x and 1.52x higher for products 4 and 8, respectively, than for product 1.
%
% If $P(x \in S)=0.3$, the increase in $P(x \in S | V(x))$ from counter product of 1 to those in values of 4 and 8 are lower and equal  the increase  1.42x and 1.52x higher when the product is 4 and 8 compared to the product of 1. 

\section{Traditional Architectures of Bloom filters with
ML models} \label{sec:traditional_architectures}
We now describe the main existing architectures that combine Bloom filters with ML models to enhance membership classification~\cite{kraska2018case,LBF} (see also~\cite{dai2020adaptive}). The two  are illustrated in Fig.~\ref{fig:combined_architectures_full}.

\begin{figure}[h!]
    \centering
    % Architecture 1
    \begin{minipage}[b]{0.38\columnwidth}
        \centering
        \resizebox{\linewidth}{!}{
            \begin{tikzpicture}[
                node distance=1.8cm,
                block/.style={rectangle, draw, minimum width=2.2cm, minimum height=0.7cm, align=center, font=\small},
                arrow/.style={-stealth, thick}
            ]
            % Nodes
            \node (oracle) [block] {Model};
            \node (bf) [block, below of=oracle] {Bloom Filter};
            
            % Entry
            \draw [arrow] (oracle.north) ++(0, 0.6) -- node[right, font=\scriptsize] {Input: element $x$} (oracle.north);
            
            % Model Outputs
            \draw [arrow] (oracle.south) -- node[right, font=\scriptsize] {negative} (bf.north);
            \draw [arrow] (oracle.east) -- ++(1.4, 0)    node (label) [pos=0.5, below, font=\scriptsize] {$f(x) \ge \tau$:}
node [below of=label, node distance=0.3cm, font=\scriptsize] {positive};

            % BF Outputs
            \draw [arrow] (bf.east) -- ++(1.4, 0) node[pos=0.5, below, font=\scriptsize] {positive};
%            \draw [arrow] (bf.east) -- ++(1.4, 0) node[pos=2.0, below, font=\scriptsize] {positive};
            \draw [arrow] (bf.south) -- ++(0, -0.6) node[below, font=\scriptsize] {negative};
            \end{tikzpicture}
        }
        \caption*{(1) Learned Bloom Filter}
    \end{minipage}
    \hfill
    % Architecture 2
    \begin{minipage}[b]{0.52\columnwidth}
        \centering
        \resizebox{\linewidth}{!}{
            \begin{tikzpicture}[
                node distance=1.8cm,
                block/.style={rectangle, draw, minimum width=2.2cm, minimum height=0.7cm, align=center, font=\small},
                arrow/.style={-stealth, thick}
            ]
            % Nodes
            \node (bf1) [block] {Bloom Filter 1};
            \node (oracle) [block, below of=bf1] {Model};
            \node (bf2) [block, below of=oracle] {Bloom Filter 2};
            
            % Entry
            \draw [arrow] (bf1.north) ++(0, 0.6) -- node[right, font=\scriptsize] {Input: element $x$} (bf1.north);
            
            % BF1 Outputs
            \draw [arrow] (bf1.west) -- ++(-1.4, 0) node[pos=0.5, below, font=\scriptsize] {negative};
            \draw [arrow] (bf1.south) -- node[right, font=\scriptsize] {positive} (oracle.north);
            
            % Model Outputs
            \draw [arrow] (oracle.south) -- node[right, font=\scriptsize] {negative} (bf2.north);
            \draw [arrow] (oracle.east) -- ++(1.4, 0) %node[pos=0.5, below, font=\scriptsize] {positive};
             node (label) [pos=0.5, below, font=\scriptsize] {$f(x) \ge \tau$:}
node [below of=label, node distance=0.3cm, font=\scriptsize] {positive};           
            % BF2 Outputs
            \draw [arrow] (bf2.east) -- ++(1.4, 0) node[pos=0.5, below, font=\scriptsize] {positive};
            \draw [arrow] (bf2.south) -- ++(0, -0.6) node[below, font=\scriptsize] {negative};
            \end{tikzpicture}
        }
        \caption*{(2) Sandwiched Learned Bloom filter}
    \end{minipage} %\vspace{-0.1in}
    \caption{Existing architectures of Bloom filters with ML models for membership classification~\cite{kraska2018case, LBF}. In each of them, the model computes a membership probability $f(x)$ for the element.}  %\vspace{-0.1in}
    \label{fig:combined_architectures_full}
\end{figure}

\textbf{Learned Bloom filter} (Kraska et al.~\cite{kraska2018case}). A model $f$ 
aims to predict the membership of an element $x$, providing a membership probability $f(x)$. With a threshold $\tau$, the architecture indicates positive if $f(x) \ge \tau$. To avoid false negatives, the Bloom filter 
  represents the set of elements whose model score is below $\tau$,
namely $\{x \in S \mid f(x)<\tau\}$.
%is used to represent all elements in the set with value of the model which is below the threshold $\tau$, namely $\{x \in S \mid f(x) < \tau\}$. 
Then if $f(x) < \tau$ the Bloom filter is queried and provides the final decision on membership. 

\textbf{Sandwiched Learned Bloom filter} (Mitzenmacher~\cite{LBF}). Two Bloom filters are used, before and after the model. The model is queried only for those elements with a positive indication of the first filter. 
Then, a positive decision is returned if either the model or the second Bloom filter outputs positive.  
In this architecture, the first filter removes most queries for non-member elements and reduces the amount of false positives of the model.

These architectures provide useful mechanisms for combining learned models with Bloom filters. In the following section, we introduce certainty-aware variants that exploit the counter information available in Counting Bloom Filters. 

%Counting Bloom Filters (CBFs) extend Bloom filters by replacing bits with counters, enabling deletions in dynamic sets. Beyond this functionality, the counter values also carry additional information that can be used to estimate the certainty of membership indications~\cite{BloomParadox}.

%\textbf{Architecture I: A learning model as a pre-filter}

\section{Enhancing Learned Models with Certainty} \label{sec:certainty_architectures}
In this section, we present four architectures that combine a learned model with Bloom filters or CBFs, while explicitly exploiting the certainty information carried by CBF counters.
Table~\ref{tab:model_choice} summarizes the main %practical considerations when choosing among the four architectures.
 considerations when choosing among the architectures.

\begin{table}[h!]
\centering
\caption{Guidelines for choosing among the     architectures.}
\label{tab:model_choice}
\small
\begin{tabular}{p{0.13\linewidth} p{0.30\linewidth} p{0.46\linewidth}}
\toprule
\textbf{Model} & \textbf{Best suited for} & \textbf{Main tradeoff} \\
\midrule
Model~1 &
Simple deployment as a drop-in replacement for a learned BF. &
Requires no front prefilter, but may have higher false-positive rate. \\
\midrule
Model~2 &
Settings where reducing false positives is more important than minimizing memory. &
Uses a front BF and a back CBF, improving filtering but requiring two structures. \\
\midrule
Model~3 &
Latency-sensitive settings with many confident model predictions. &
Allows early positive decisions, reducing CBF queries, but inherits the memory cost of Model~2. \\
\midrule
Model~4 &
Memory-constrained settings where tight integration between learning and filtering is acceptable. &
Reuses one CBF for screening and posterior decisions, but creates statistical dependence between stages. \\
\bottomrule
\end{tabular}
\end{table}

\subsection{Notation}
Let $f(x)\in[0,1]$ denote the score produced by the learned oracle for an element $x$, interpreted as an estimate of the membership probability of $x$ in the set $S$. Let $\tau\in[0,1]$ denote the oracle decision threshold: if $f(x)\ge\tau$, the oracle predicts a positive membership decision; otherwise, it predicts negative.\footnote{In this work we treat $\tau$ as fixed. In future work, it may be interesting to adapt $\tau$ depending on the downstream stages of the architecture, e.g., making the oracle more conservative when additional filtering stages are available.} Table~\ref{tab:notation} summarizes the main notation used throughout the paper.

\begin{table}[t!]
\centering 
\caption{Notation.}  \vspace{-0.1in}
\label{tab:notation}
\resizebox{\columnwidth}{!}{%
\begin{tabular}{l l}
\toprule
Symbol & Description \\
\midrule
$f(x)$ & score produced by the learned oracle \\
$V(x)$ & vector of counters associated with $x$ in the CBF \\
$\Pi(x)$ & counter product $\prod_{j=1}^{k} c_j$ \\
$\prior$ & prior membership probability \\
$\prior_0$ & baseline prior $P(x\in S)$ \\
$\prior_1$ & learned prior $f(x)$ \\
$\prior_2$ & MAP-style prior $f(x,V(x))$ \\
$h_{\prior}(x)$ & CBF posterior decision rule using prior $\prior$ \\ 
$\varepsilon_f$ & false-positive rate of the learned oracle \\
$\varepsilon_B$ & false-positive rate of the front BF \\
$\varepsilon_{\mathrm{CBF}}$ & false-positive rate of the CBF when used as prefilter \\
\bottomrule
\end{tabular} } \vspace{-0.2in}
\end{table}  

Let $V(x)=(c_1,\dots,c_k)$ be the vector of counters associated with $x$ in the Counting Bloom Filter (CBF), where $k$ is the number of hash functions. We define the counter product $\Pi(x)=\prod_{j=1}^{k} c_j$. 

Following~\cite{BloomParadox}, when all counters are nonzero the CBF induces a posterior membership probability of the form
\begin{equation}
P(x\in S \mid V(x); \prior)
=
\frac{m^k \cdot \Pi(x) \cdot \prior}
{m^k \cdot \Pi(x) \cdot \prior + (n\cdot k)^k \cdot (1-\prior)},
\label{eq:cbf_posterior}
\end{equation}
where $\prior$ denotes the prior membership probability used in the posterior computation.    

A subtle point is that the prior used in the CBF posterior rule depends on the
reference population. If the CBF is meant to be oracle-agnostic and reusable
across different learned oracles, then $\prior$ can be estimated over the full query
universe. If the CBF is specialized to a particular oracle and architecture,
then $\prior$ should instead be estimated over the conditional universe of queries
that reach the CBF stage, such as those satisfying $f(x)<\tau$.
The latter choice may improve calibration, but couples the CBF posterior rule to
a specific oracle.

We consider three possible priors:
\[
\prior_0 = P(x\in S), \qquad
\prior_1 = f(x), \qquad
\prior_2 = f(x,V(x)),
\]
where $p_0$ is a population prior, $p_1$ is the oracle prior,
and $p_2$ is a MAP-style prior that depends on both $x$ and the counters.
For the false-positive-rate analysis, 
let $\varepsilon_f$ denote the false-positive rate of the learned oracle.  Let $\varepsilon_B$ and $\varepsilon_{\mathrm{CBF}}$  denote the false-positive rates of the front
BF and CBF prefilters, respectively.
%
% For the false-positive-rate analysis, let $\varepsilon_f$ denote the false-positive rate of the learned oracle and let $\varepsilon_B$ denote the false-positive rate of the front Bloom filter. 
%
%
%Finally, let $h_p(x)$ denote the CBF posterior decision rule that outputs  $1$ when $P(x \in S \mid V(x); p) \ge \eta$, where $\eta \in [0,1]$ is a posterior decision threshold.
%
Finally, let $h_p(x)$ denote the CBF posterior decision rule that outputs $1$
when all counters associated with $x$ are nonzero and
$P(x \in S \mid V(x);p) \ge \eta$, where $\eta \in [0,1]$ is the posterior
decision threshold, e.g., $\eta={1}/({\alpha+1})$ under the Bayes-optimal rule from Property~1. 
\begin{equation} h_p(x)=1 \quad\Longleftrightarrow\quad c_j>0\ \forall j \ \text{ and }\ P(x\in S\mid V(x);p)\ge \eta. \label{eq:hpx1} \end{equation}
Unlike a standard Bloom-filter membership test, the posterior rule $h_p(x)$ may
produce false negatives when the posterior threshold $\eta$ rejects a true member.
Thus, the architectures below trade false positives against false negatives.

The four architectures described in the following sections are illustrated in Fig.~\ref{fig:models_1_to_4} and differ in how they combine three sources of information: the learned score $f(x)$, the CBF counter vector $V(x)$, and an optional front-end prefilter. Model~1 uses no prefilter, Models~2 and~3 include a front filter, and Model~4 reuses the same CBF both as a prefilter and as a certainty-aware decision module.

\begin{comment}
\begin{figure*}[t!]
    \centering

    \begin{minipage}[b]{0.24\textwidth}
        \centering
        \includegraphics[width=\linewidth]{model1b-crop.pdf}
        \caption*{(1) Model 1}
    \end{minipage}
    \hfill
    \begin{minipage}[b]{0.24\textwidth}
        \centering
        \includegraphics[width=\linewidth]{model2b-crop.pdf}
        \caption*{(2) Model 2}
    \end{minipage}
    \hfill
    \begin{minipage}[b]{0.24\textwidth}
        \centering
        \includegraphics[width=\linewidth]{model3b-crop.pdf}
        \caption*{(3) Model 3}
    \end{minipage}
    \hfill
    \begin{minipage}[b]{0.24\textwidth}
        \centering
        \includegraphics[width=\linewidth]{model4b-crop.pdf}
        \caption*{(4) Model 4}
    \end{minipage}
%\vspace{-0.1in}
    \caption{Architectures combining learned membership models with Bloom filters and Counting Bloom Filters. }%\ori{I printed it, resolution seems fine but not great - can you imrpove resoultion of the text?}}
    \label{fig:models_1_to_4} %\vspace{-0.1in}
\end{figure*}

\end{comment}

 \begin{figure*}[t]
    \centering
    \scriptsize

    %%%%%%%%%%%%%%%%%%%%%%%%%%%%%%
    % Model 1
    %%%%%%%%%%%%%%%%%%%%%%%%%%%%%%
    \begin{minipage}[t]{0.24\textwidth}
        \centering
        \resizebox{\linewidth}{!}{
            \begin{tikzpicture}[
                node distance=1.25cm,
                block/.style={
                    rectangle,
                    draw,
                    minimum width=1.9cm,
                    minimum height=0.55cm,
                    align=center,
                    font=\scriptsize
                },
                arrow/.style={-stealth, thick}
            ]
            \node (oracle) [block] {Learned\\Oracle};
            \node (cbf) [block, below of=oracle] {CBF $C_1$\\using $p_0$ as prior};

            \draw [arrow] (oracle.north) ++(0,0.45)
                -- node[right, font=\scriptsize] {Input: element $x$} (oracle.north);

            \draw [arrow] (oracle.east) -- ++(1.0,0)
                node[pos=0.5, above, font=\scriptsize] {$f(x)\ge\tau$}
                node[pos=0.5, below, font=\scriptsize] {positive};

            \draw [arrow] (oracle.south) --
                node[right, font=\scriptsize] {$f(x)<\tau$}
                node[left, font=\scriptsize] {$p_0=P(x\in S)$}
                (cbf.north);

            \draw [arrow] (cbf.east) -- ++(1.0,0)
                node[pos=0.5, below, font=\scriptsize] {positive};

            \draw [arrow] (cbf.west) -- ++(-1.0,0)
                node[pos=0.5, below, font=\scriptsize] {negative};
            \end{tikzpicture}
        }

        \caption*{(1) Model 1}
    \end{minipage}
    \hfill
    %%%%%%%%%%%%%%%%%%%%%%%%%%%%%%
    % Model 2
    %%%%%%%%%%%%%%%%%%%%%%%%%%%%%%
    \begin{minipage}[t]{0.24\textwidth}
        \centering
        \resizebox{\linewidth}{!}{
            \begin{tikzpicture}[
                node distance=1.25cm,
                block/.style={
                    rectangle,
                    draw,
                    minimum width=1.9cm,
                    minimum height=0.55cm,
                    align=center,
                    font=\scriptsize
                },
                arrow/.style={-stealth, thick}
            ]
            \node (bf) [block] {BF $B_1$\\with all $S$};
            \node (oracle) [block, below of=bf] {Learned\\Oracle};
            \node (cbf) [block, below of=oracle] {CBF $C_1$\\using $p_1$ as prior};

            \draw [arrow] (bf.north) ++(0,0.45)
                -- node[right, font=\scriptsize] {Input: element $x$} (bf.north);

            \draw [arrow] (bf.west) -- ++(-1.0,0)
                node[pos=0.5, below, font=\scriptsize] {negative};

            \draw [arrow] (bf.south) -- (oracle.north);

            \draw [arrow] (oracle.south) --
                node[left, font=\scriptsize] {$p_1=f(x)$}
                (cbf.north);

            \draw [arrow] (cbf.east) -- ++(1.0,0)
                node[pos=0.5, below, font=\scriptsize] {positive};

            \draw [arrow] (cbf.west) -- ++(-1.0,0)
                node[pos=0.5, below, font=\scriptsize] {negative};
            \end{tikzpicture}
        }

        \caption*{(2) Model 2}
    \end{minipage}
    \hfill
    %%%%%%%%%%%%%%%%%%%%%%%%%%%%%%
    % Model 3
    %%%%%%%%%%%%%%%%%%%%%%%%%%%%%%
    \begin{minipage}[t]{0.24\textwidth}
        \centering
        \resizebox{\linewidth}{!}{
            \begin{tikzpicture}[
                node distance=1.25cm,
                block/.style={
                    rectangle,
                    draw,
                    minimum width=1.9cm,
                    minimum height=0.55cm,
                    align=center,
                    font=\scriptsize
                },
                arrow/.style={-stealth, thick}
            ]
            \node (bf) [block] {BF $B_1$\\with all $S$};
            \node (oracle) [block, below of=bf] {Learned\\Oracle};
            \node (cbf) [block, below of=oracle] {CBF $C_1$\\using $p_1$ as prior};

            \draw [arrow] (bf.north) ++(0,0.45)
                -- node[right, font=\scriptsize] {Input: element $x$} (bf.north);

            \draw [arrow] (bf.west) -- ++(-1.0,0)
                node[pos=0.5, below, font=\scriptsize] {negative};

            \draw [arrow] (bf.south) -- (oracle.north);

            \draw [arrow] (oracle.east) -- ++(1.0,0)
                node[pos=0.5, above, font=\scriptsize] {$f(x)\ge\tau$}
                node[pos=0.5, below, font=\scriptsize] {positive};

            \draw [arrow] (oracle.south) --
                node[right, font=\scriptsize] {$f(x)<\tau$}
                node[left, font=\scriptsize] {$p_1=f(x)$}
                (cbf.north);

            \draw [arrow] (cbf.east) -- ++(1.0,0)
                node[pos=0.5, below, font=\scriptsize] {positive};

            \draw [arrow] (cbf.west) -- ++(-1.0,0)
                node[pos=0.5, below, font=\scriptsize] {negative};
            \end{tikzpicture}
        }

        \caption*{(3) Model 3}
    \end{minipage}
    \hfill
    %%%%%%%%%%%%%%%%%%%%%%%%%%%%%%
    % Model 4
    %%%%%%%%%%%%%%%%%%%%%%%%%%%%%%
    \begin{minipage}[t]{0.24\textwidth}
        \centering
        \resizebox{\linewidth}{!}{
            \begin{tikzpicture}[
                node distance=1.25cm,
                block/.style={
                    rectangle,
                    draw,
                    minimum width=1.9cm,
                    minimum height=0.55cm,
                    align=center,
                    font=\scriptsize
                },
                arrow/.style={-stealth, thick}
            ]
            \node (screen) [block] {CBF $C_1$\\with all $S$};
            \node (oracle) [block, below of=screen] {Learned MAP\\Oracle};
            \node (posterior) [block, below of=oracle] {Same CBF $C_1$\\using $p_2$ as prior};

            \draw [arrow] (screen.north) ++(0,0.45)
                -- node[right, font=\scriptsize] {Input: element $x$} (screen.north);

            \draw [arrow] (screen.west) -- ++(-1.0,0)
                node[pos=0.5, above, font=\scriptsize] {$\exists j{:}c_j=0$}
                node[pos=0.5, below, font=\scriptsize] {negative};

            \draw [arrow] (screen.south) --
                node[left, font=\scriptsize] {$x,V(x)$}
                node[right, font=\scriptsize] {$c_j>0\ \forall j$}
                (oracle.north);

            \draw [arrow] (oracle.east) -- ++(1.0,0)
             node[pos=0.4, above, font=\scriptsize, align=center]
  %  {$f(x,V(x)) $\\$ \;\; \qquad \ge\tau$}
    {$f(x,V(x))  $\\$ \ge\tau$}
                node[pos=0.5, below, font=\scriptsize] {positive};

            \draw [arrow] (oracle.south) --
                node[right, font=\scriptsize] {$f(x,V(x))<\tau$}
                node[left, font=\scriptsize] {$p_2=f(x,V(x))$}
                (posterior.north);

            \draw [arrow] (posterior.east) -- ++(1.0,0)
                node[pos=0.5, below, font=\scriptsize] {positive};

            \draw [arrow] (posterior.west) -- ++(-1.0,0)
                node[pos=0.5, below, font=\scriptsize] {negative};
            \end{tikzpicture}
        }

        \caption*{(4) Model 4}
    \end{minipage} \vspace{-0.1in}
\caption{Architectures combining learned membership models with Bloom filters and Counting Bloom Filters. In vertical arrows, labels on the right indicate the routing condition, while labels on the left indicate the information passed to the next stage, such as the prior used by the CBF posterior test. Model~1 uses the population prior $p_0=P(x\in S)$. Models~2 and~3 use the learned prior $p_1=f(x)$, while Model~4 uses the MAP-style prior $p_2=f(x,V(x))$.}
    \label{fig:models_1_to_4}
\end{figure*}

\subsection{Model 1: Learned Counting Bloom Filter}
Model~1 is the direct certainty-aware counterpart of the learned Bloom filter. The learned model is queried first. If the model is sufficiently confident that $x$ belongs to the set, namely if $f(x)\ge \tau$,
 the system outputs a positive decision immediately. Otherwise, the query is forwarded to a CBF.

The CBF is then used not merely as a binary backup structure, but as a posterior estimator. If at least one of the $k$ counters is zero, the output is negative. Otherwise, the decision is based on the posterior probability $P(x\in S \mid V(x);\prior)$ computed using~\eqref{eq:cbf_posterior}, 
where the prior is taken to be $\prior_0=P(x\in S)$, e.g., a population-level estimate.  
Hence, the decision rule is
\begin{equation}
\hat{y}_1(x)=
\begin{cases}
1, & f(x)\ge \tau,\\[3pt]
1, & f(x)<\tau,\; c_j>0\ \forall j,\; 
P(x\in S\mid V(x);\prior_0)\ge \eta,\\[3pt]
0, & \text{otherwise},
\end{cases}
  \nonumber
\end{equation}
where $\eta$ is a posterior threshold.

Model~1 preserves the role of the learned model as a front-end classifier, but replaces the standard Bloom filter fallback by a certainty-aware structure. This allows the second stage to distinguish weak positive indications from strong ones, which a binary Bloom filter cannot do.

\emph{False-positive rate. }
For a non-member key, a false positive occurs either because the learned oracle already outputs positive, or because the oracle outputs negative and the CBF posterior test accepts the key. Thus,  
\begin{equation}
\mathrm{FPR}_1
=
\underbrace{\varepsilon_f}_{\text{oracle FP}}
+
(1-\varepsilon_f) \cdot  \underbrace{
\Pr\!\left(
h_{\prior_0}(x)=1
\;\middle|\;
x\notin S,  f(x)<\tau 
\right)
}_{\text{CBF accepts a non-member}}.
\label{eq:fpr_model1}
\end{equation}
This is the only model without a front prefilter.  

\emph{Discussion. }
Model~1 is the most direct certainty-aware extension of a learned Bloom-filter pipeline.  First, there is \emph{no pre-filter}, which tends to produce a relatively high false-positive rate. Second, for low prior values, using the CBF membership probability is exactly where the CBF is most useful, because the product of the counters carries proportionally more weight in the posterior decision.  

\subsection{Model 2: Asymmetric Sandwich}
Model~2 adds a standard Bloom filter $B_1$ before the learned model. The first-stage Bloom filter filters out a large fraction of non-members before they reach the learned model. Only if $B_1(x)=1$ is the query passed to the learned model.

In this architecture, the learned model does not directly produce the final binary output. Instead, its score $f(x)$ is
used as the prior in the posterior computation performed by the CBF stage. 
Thus, conditioned on $B_1(x)=1$ and on all CBF counters being positive, the posterior probability is obtained from~\eqref{eq:cbf_posterior} by setting the prior to be $\prior_1 = f(x)$. The resulting decision rule is
\begin{equation}
\hat{y}_2(x)=
\begin{cases}
0, & B_1(x)=0,\\[3pt]
1, & B_1(x)=1,\; c_j>0\ \forall j,\; P(x\in S\mid V(x);\prior_1)\ge \eta,\\[3pt]
0, & \text{otherwise}.
\end{cases}  \nonumber
\end{equation}

Model~2 is asymmetric because the first filter is a standard Bloom filter and the second is a CBF. The key idea is that the learned model contributes soft prior information, while the CBF contributes structural evidence from the counters. Their combination yields a posterior that is typically more informative than either source alone.

\emph{False-positive rate. }
The prefilter reduces the false-positive rate by the front-filter term. Writing that explicitly,
\begin{equation}
\mathrm{FPR}_2
=
\varepsilon_B \cdot
\Pr\!\left(
h_{\prior_1}(x)=1
\;\middle|\;
x\notin S,  B_1(x)=1
\right).
\label{eq:fpr_model2}
\end{equation}
Here the oracle output $f(x)$ serves as the prior $\prior$ in the posterior computation $P(x\in S \mid V(x);\prior)$ performed by the CBF stage.

\emph{Discussion. }
This model is asymmetric because the front stage is a standard Bloom filter while the back stage is a CBF. We have two benefits: the prefilter reduces the false-positive rate, and the oracle is used as a prior input to the CBF, which makes the final decision more informative than a binary Bloom-filter test alone.  

\subsection{Model 3: Asymmetric Sandwich with Early Decision}
Model~3 is similar to Model~2, but allows the learned model to make an early positive decision. As before, a first Bloom filter $B_1$ is queried. If $B_1(x)=0$, the final decision is negative. If $B_1(x)=1$, the query reaches the learned model. 

This architecture has an additional advantage: the CBF stage is invoked primarily for elements whose score $f(x)$ is below the threshold $\tau$. In this regime the prior membership probability is relatively small, and the counter product $\Pi(x)$ has a stronger influence on the posterior probability (see Fig.~\ref{figure_counter_product_stats}(c)). Thus, the CBF posterior test can be particularly informative for such low-score elements.

If the model score is above threshold, i.e., 
$f(x)\ge \tau$,
the system outputs a positive decision immediately. If instead $f(x)<\tau$, the query is passed to a CBF, where the posterior probability $P(x\in S \mid V(x);\prior_1)$ uses prior $\prior_1=f(x)$.
%
%
% The posterior probability $P(x\in S \mid V(x))$ is then computed using~\eqref{eq:cbf_posterior} with prior $\prior = P(x\in S)$, replacing the generic prior $p$ by $P(x\in S)$ and the counter product $\Pi(x)$ by $\prod_{j=1}^{k} c_j$.
 %
%
Thus,  
\begin{equation}
\hat{y}_3(x)=
\begin{cases}
0, & B_1(x)=0,\\[3pt]
1, & B_1(x)=1,\; f(x)\ge \tau,\\[3pt]
1, & B_1(x)=1,\; f(x)<\tau,\; c_j>0\ \forall j,\; \\
&P(x\in S\mid V(x);\prior_1)\ge \eta,\\[3pt]
0, & \text{otherwise}.
\end{cases}
\label{eq:model3-rule} \nonumber
\end{equation}

Compared to Model~2, this architecture can reduce latency and computational cost by skipping the CBF stage when the learned model is already highly confident. At the same time, low-confidence cases still benefit from the certainty-aware reasoning offered by the CBF.

\emph{False-positive rate. }
A false positive in Model~3 occurs when a non-member key  passes
the front Bloom filter, and  either the oracle produces a false
positive or, after a negative oracle output, the CBF posterior test
accepts the key. This yields
\begin{equation}
\begin{aligned}
\mathrm{FPR}_3
&=
\varepsilon_B \cdot
\Bigg(
\varepsilon_{f \mid B}
+
(1-\varepsilon_{f \mid B}) \cdot
\Pr\!\left(
h_{\prior_1}(x)=1
\;\middle|\;
\substack{
x\notin S,\ f(x)<\tau,\\
B_1(x)=1
}
\right)
\Bigg),
\end{aligned}
\label{eq:fpr_model3}
\end{equation}
where $\varepsilon_{f \mid B} = \Pr\!\left(
f(x)\ge \tau
\mid
x\notin S,\ B_1(x)=1
\right)$.
%\ori{can you add $\cdot$ signs to the last equation? eg after the first component but also later}

\emph{Discussion. } 
Model 3 extends Model 2 by allowing an early positive decision when the learned oracle is sufficiently confident, while still using the oracle score as prior in the fallback CBF stage. 
The prefilter reduces the false-positive rate, and the same low-prior CBF advantage from Model~1 is retained in the fallback stage. Relative to Model~2, Model~3 also has an early decision path, which can reduce computation when the oracle is sufficiently confident.  

\subsection{Model 4: Symmetric MAP Sandwich}
Model~4 provides the tightest integration between learning and the counting filter. A single CBF is used both as an initial screening device and as a final posterior-based decision device. First, the input $x$ is checked against the CBF in the classical sense: if at least one associated counter is zero, the output is immediately negative. Otherwise, the counter vector $V(x)=(c_1,\dots,c_k)$ is extracted and passed, together with $x$, to a learned model.  Algorithm~\ref{algo:algo4} summarizes Model~4.

\SetAlgoNoEnd
\begin{algorithm}[h!]
\DontPrintSemicolon  
\caption{Model 4: Symmetric MAP Sandwich} \label{algo:algo4}
\KwIn{Element $x$, CBF, MAP oracle $f(x,V(x))$,  threshold   $\tau$}
\KwOut{Positive / Negative }
Read counters $V(x)=\{c_i\}_{i=1}^k$ of $x$ from the CBF\;
\If{there exists $i$ such that $c_i=0$}{
    \Return{\texttt{Negative}} \tcp*{BF-like prefilter}
}
\If{$f(x, V(x))\ge \tau$}{
    \Return{\texttt{Positive}}\;
}
Use $\prior_2=f(x,V(x))$ as an effective MAP-style prior in the CBF posterior test\;
\eIf{$P(x\in S\mid V(x);\prior_2)\ge  \eta
$}{    \Return{\texttt{Positive}}\;
}{
    \Return{\texttt{Negative}}\;
} 
\end{algorithm}

Unlike the previous models, the learned model in Model~4 may explicitly use both the feature representation of $x$ and the counter information. Denote its output by $f(x,V(x))$. If $f(x,V(x))\ge \tau$, the architecture outputs a positive decision. 
Otherwise, the same score is used as an effective MAP-style prior in the final posterior computation. 
% Otherwise, the same score is used as the prior in the final posterior computation. 
In particular, $P(x\in S \mid V(x);\prior_2)$ is obtained from~\eqref{eq:cbf_posterior} by setting $\prior_2 = f(x,V(x))$.

The decision rule is
\begin{align}
&\hat{y}_4(x)= \nonumber \\
&
\begin{cases}
0, & \exists j \text{ such that } c_j=0,\\[3pt]
1, & c_j>0\ \forall j,\; f(x,V(x))\ge \tau,\\[3pt]
1, & c_j>0\ \forall j,\; f(x,V(x))<\tau,  P(x\in S\mid V(x);\prior_2)\ge \eta,\\[3pt]
0, & \text{otherwise}.
\end{cases} 
\label{eq:model4-rule} \nonumber
\end{align}

Model~4 is symmetric because the same CBF appears on both sides of the learned component. It is also the most flexible architecture, since the learned model can exploit the full counter vector rather than just a binary Bloom-filter output. This allows the model to learn richer decision rules that combine content-based features of $x$ with the certainty signal carried by the filter.

 \emph{False-positive rate.}
Let \(C_1(x)=1\) denote the event that the CBF screening stage
returns a positive indication, i.e., all counters associated with \(x\)
are nonzero. A false positive in Model~4 occurs when a non-member
key passes the CBF screening stage and the MAP oracle produces an
early positive decision, or when the CBF posterior test accepts the
key after the oracle output is below threshold.
Thus, by the definition of $h_p$ in~\eqref{eq:hpx1}, $h_{\prior_2}(x)=1 \Rightarrow C_1(x)=1$, and
%
%Thus, noting that \(h_{\prior_2}(x)=1 \Rightarrow C_1(x)=1\),
\begin{equation}
\begin{aligned}
\mathrm{FPR}_4
&=
\Pr\!\left(
C_1(x)=1,\ f(x,V(x))\ge\tau
\mid x\notin S
\right) \\
&\quad+
\Pr\!\left(
h_{\prior_2}(x)=1,\ f(x,V(x))<\tau
\mid x\notin S
\right).
\end{aligned}
\label{eq:fpr_model4_exact}
\end{equation}
Equivalently, writing $\varepsilon_{\mathrm{CBF}}
=
\Pr\!\left(C_1(x)=1\mid x\notin S\right)$
and
$\varepsilon_{f\mid CBF}
=
\Pr\!\left(
f(x,V(x))\ge\tau
\mid x\notin S,\ C_1(x)=1
\right)$, 
we obtain
\begin{equation}
\begin{aligned}
\mathrm{FPR}_4
&=
\varepsilon_{\mathrm{CBF}} \cdot \varepsilon_{f\mid CBF}
+
\Pr\!\left(
h_{\prior_2}(x)=1,\ f(x,V(x))<\tau
\mid x\notin S
\right).
\end{aligned} \nonumber
\end{equation}

\emph{Discussion. }
Model~4 forms a symmetric architecture in which the same CBF structure
appears both before and after the learned oracle. This design provides
several advantages. First, the prefilter stage reduces the number of
queries reaching the learned model, 
which can reduce the overall false-positive rate. %thereby lowering the overall false positive rate.
Second, reusing the same CBF for both stages reduces
memory requirements compared to architectures that maintain separate
filters. Third, the MAP oracle uses both the element features and the
CBF counters to produce an informed prior $\prior_2$. Finally, the posterior
CBF decision continues to exploit the strong certainty signal provided
by the counters, which is particularly beneficial 
%when the baseline membership prior is small.
for a small baseline membership prior. 

\subsection{Discussion}
The four architectures differ in how they combine three sources of information: a binary filter output, a learned score, and the CBF-derived certainty signal. Model~1 is the simplest certainty-aware extension of the learned Bloom filter. Model~2 uses the learned score as a prior inside the CBF posterior. Model~3 adds an early-exit mechanism based on confident model predictions. Model~4 allows the strongest interaction between learning and filtering by feeding counter information directly into the learned model itself.

These architectures also induce different operating points in terms of false positives, false negatives, memory footprint, and inference cost.  With respect to false negatives, although the underlying BF/CBF structures have no false negatives, the complete architectures may have false negatives
because the learned oracle and/or the posterior CBF decision rule can reject true
members. Hence the proposed schemes should be viewed as operating
points in a false-positive/false-negative tradeoff, rather than as strict
BF replacements with one-sided error.

In particular, the architectures differ in how many filter structures are required and whether a CBF must be maintained in addition to a standard BF. 
Model~1 is likely to be easiest to deploy as a drop-in replacement for a learned Bloom filter, while Model~4 offers greater flexibility at the price of tighter coupling between the model and the data structure.

\section{Conclusion}
\label{sec:dynamic_updates}

Counting Bloom Filters (CBFs) are versatile data structures for compact set representation, especially when deletions or updates are required.
Beyond supporting deletions, their counters expose information that can be
used to estimate posterior membership probabilities and quantify the certainty
of positive indications. We showed how this certainty signal can be combined
with learned models through four architectures integrating Bloom filters,
CBFs, and learned predictors.

These architectures differ in how they combine prefiltering, learned priors,
early decisions, and certainty-aware posterior tests, which may lead to
different operating points in terms of false-positive rate, inference cost,
and memory usage. A practical challenge arises when the represented set changes over time: while the CBF can adapt   through insertions and deletions, the learned model typically remains fixed until retraining. In practice, this suggests a hybrid update strategy in which the CBF is updated online, while the learned oracle is retrained periodically or when performance degradation is detected.

Several directions remain for future work. First, it would be interesting to
study how to optimally allocate memory across different components of the
architecture, such as front filters, counting filters, and learned models.
Second, the decision threshold $\tau$ used by the learned oracle could
potentially be adapted depending on the downstream stages of the architecture,
allowing earlier decisions to become more conservative when additional
filtering stages are available. Third, it may be worthwhile to explore other
probabilistic data structures that, like CBFs, expose internal signals that
can be interpreted as measures of certainty rather than only binary membership
decisions.
%\input{background}

%\input{method}

%\input{evaluation}

%\input{conclusion}

\bibliographystyle{ACM-Reference-Format}
\bibliography{ref}

\end{document}